\documentclass[11pt,draftcls]{IEEEtran}
\onecolumn
\usepackage{url} 
\usepackage{lineno}
\usepackage{color}

\usepackage{algorithm,algorithmic}
\usepackage{verbatim}
\usepackage{setspace}

\usepackage{cite}
\usepackage[dvipsnames,table]{xcolor}
\usepackage{flushend}

\usepackage[cmex10]{amsmath}

\usepackage{array}
\usepackage{graphicx}
\usepackage{subfigure}
\usepackage{booktabs}
\usepackage{multirow}
\usepackage[export]{adjustbox}

\newcolumntype{x}[1]{%
>{\centering\hspace{0pt}}p{#1}}%

\begin{document}

\noindent\rule{\textwidth}{1pt}
This manuscriptis a preprintand has been submitted for publication in IEEE Transactions on Geoscience and Remote Sensing. Please note that, despite having undergone peer-review, the manuscript has yet to be formally accepted for publication. Subsequent versions of this manuscript may have slightly different content.If accepted, the final version of this manuscript will be available via the ‘Peer-reviewed Publication DOI’ link on the right-hand side of this webpage. Please feel free to contact any of the authors; we welcome feedback.

\vspace{1cm}
\noindent Authors:

\noindent N. Anantrasirichai, Visual Information Laboratory, University of Bristol, UK ( N.Anantrasirichai@bristol.ac.uk)

\noindent  J. Biggs, School of Earth Sciences, University of Bristol, UK (Juliet.Biggs@bristol.ac.uk)

\noindent  K. Kelevitz, COMET, School of Earth and Environment, University of Leeds, UK (k.kelevitz@leeds.ac.uk)

\noindent  Z. Sadeghi, COMET, School of Earth and Environment, University of Leeds, UK (z.sadeghi@leeds.ac.uk)

\noindent  T .Wright, COMET, School of Earth and Environment, University of Leeds, UK (t.j.wright@leeds.ac.uk)

\noindent  J. Thompson, School of Earth and Environment, University of Leeds, UK (j.thompson@leeds.ac.uk)

\noindent  A. Achim, Visual Information Laboratory, University of Bristol, UK (Alin.Achim@bristol.ac.uk)

\noindent  D. Bull, Visual Information Laboratory, University of Bristol, UK (Dave.Bull@bristol.ac.uk)

\noindent\rule{\textwidth}{1pt}

\newpage
\title{Deep Learning Framework for Detecting Ground Deformation in the Built Environment using Satellite InSAR data}

\author{Nantheera Anantrasirichai,~\IEEEmembership{Member,~IEEE,}
        Juliet~Biggs, Krisztina~Kelevitz, Zahra~Sadeghi, Tim~Wright, James~Thompson, Alin~Achim,~\IEEEmembership{Senior Member,~IEEE} 
        and~David~Bull,~\IEEEmembership{Fellow,~IEEE}% <-this % stops a space
\thanks{This work was supported by the Natural Environment Research Council Digital Environment Programme Feasibility Study (NE/S016104/1)}
\thanks{N. Anantrasirichai, A. Achim and D. Bull are with the Visual Information Laboratory, University of Bristol, UK (email: N.Anantrasirichai@bristol.ac.uk, alin.achim@bristol.ac.uk, Dave.Bull@bristol.ac.uk).}
\thanks{J. Biggs is with the School of Earth Sciences, University of Bristol, UK (email: Juliet.Biggs@bristol.ac.uk).}
\thanks{K. Kelevitz, Z. Sadeghi, T. Wright and J. Thompson are with the COMET, School of Earth and Environment, University of Leeds, UK (email: k.kelevitz@leeds.ac.uk, z.sadeghi@leeds.ac.uk, t.j.wright@leeds.ac.uk).}
\thanks{Manuscript received xx, 2020; revised xx, 20xx.}}


%\markboth{Submit to IEEE Transactions on Geoscience and Remote Sensing}%
%{N. Anantrasirichai \MakeLowercase{\textit{et al.}}: Bare Demo of IEEEtran.cls for IEEE Journals}


\maketitle

\begin{abstract}
The large volumes of Sentinel-1 data produced over Europe are being used to develop pan-national ground motion services. However, simple analysis techniques like thresholding cannot detect and classify complex deformation signals reliably making providing usable information to a broad range of non-expert stakeholders a challenge. Here we explore the applicability of deep learning approaches by adapting a pre-trained convolutional neural network (CNN) to detect deformation in a national-scale velocity field. For our proof-of-concept, we focus on the UK where previously identified deformation is associated with coal-mining, ground water withdrawal, landslides and tunnelling. The sparsity of measurement points and the presence of spike noise make this a challenging application for deep learning networks, which involve calculations of the spatial convolution between images. Moreover, insufficient ground truth data exists to construct a balanced training data set, and the deformation signals are slower and more localised than in previous applications.  We propose three enhancement methods to tackle these problems: i) spatial interpolation with modified matrix completion, ii) a synthetic training dataset based on the characteristics of real UK velocity map, and iii) enhanced over-wrapping techniques.  Using velocity maps spanning 2015-2019, our framework detects several areas of coal mining subsidence, uplift due to dewatering, slate quarries, landslides and tunnel engineering works. The results demonstrate the potential applicability of the proposed framework to the development of automated ground motion analysis systems.
\end{abstract}


% \fi
\IEEEpeerreviewmaketitle

\section{Introduction}

For the last few decades, it has been possible to accurately measure ground deformation from space using Interferometric Synthetic Aperture Radar (InSAR) \cite{Massonnet:Radar:1998}. Recent advances in processing techniques and computing power (e.g. \cite{Spaans:InSAR:2016}), coupled with the launch of the Sentinel-1 satellites have laid the foundation for millimetre-scale monitoring of ground deformation across Europe in near real time. This has obvious potential for monitoring ground movement in urban and semi-rural environments. We use the United Kingdom as a test case, where the average annual cost to the insurance industry of ground motion is estimated to be over {\pounds}250M \cite{plante2012review,pritchard2013soil}. Incidents affecting critical infrastructure, such as mainline railways or dams, can be associated with multi-million pound costs, even for a single slope failure event. The sources of deformation are both natural and anthropogenic: subsidence and heave due to the legacy of the coal mining and quarrying industries \cite{McCay:meta:2018},  shrink and swell of shallow clays \cite{Aldiss:geo:2014}, natural sinkholes \cite{Gutierrez:identification:2008}, landslides \cite{Chambers:three:2011}, coastal erosion \cite{Whiteley:Geophysical:2019}, and engineering work, such as tunnelling \cite{Milillo:multi:2018}. 

The Sentinel-1 satellites acquire data over a 250-km swath at a 4 m by 14 m spatial resolution every 6 days on both ascending and descending tracks, generating a large quantity of data. So far, efforts have largely focused on improving data processing methods and capacity \cite{Licsar:2016}, but the need for manual inspection and expert interpretation are also barriers to the timely dissemination of information. Various approaches to automatic detection have been tested, for example, the authors in \cite{Raspini:continous:2018} use a threshold of 10mm/yr to identify anomalies in time-series data from Northern Italy. However, applying a threshold in the spatial domain is not reliable due to the effect of reference-point selection and the performance deteriorates heavily for noisy and low coherence signals. 
Albino et. al. \cite{Albino:automated:2020} used receiver operating characteristics to demonstrate that applying a cumulative sum control chart  \cite{PAGE:CUSUM:1954} to the time-series improves detection performance over simple thresholding.  However, both these methods work on individual pixels and do not take into account the high spatial resolution  information that is a major advantage for InSAR.  Independent Component Analysis (ICA) has been used to separate deformation from noise based on the assumptions that the signals are statistically independent and non-Gaussian \cite{ebmeier2016,Chaussard:Remote:2017, Gaddes:blind:2018}. However, the main drawback for the use of ICA in automated systems is the uncertainty in the order of the separated components, known as the permutation problem \cite{Sawada:rebust:2004}.

In this paper, we employ a convolutional neural network (CNN) to automatically detect ground deformation across the United Kingdom. The CNN models the spatial characteristics of InSAR data and then recognises the difference between deformation and atmospheric noise. We base our study on a transferable machine learning approach that has already been successfully used for detecting volcanic deformation in global InSAR data \cite{Anantrasirichai:Application:2018, Anantrasirichai:deep:2019, Anantrasirichai:application:2019}. 
Adapting these approaches for detecting urban deformation is conceptually straightforward, but challenging to implement due to the unsuitable nature of available signals for CNN-based algorithms. The sources of deformation in the UK are much shallower and slower than in volcanic environments, meaning the deformation has a smaller magnitude and spatial extent. The spatially variable coherence and associated processing methods means that InSAR data for the UK is typically sparse and has different noise characteristics to volcanic environments.

In this paper, we  propose three novel contributions to address these problems:  i) spatial interpolation with a modified  matrix completion method to tackle sparsity and simultaneously mitigate noise due to atmospheric effects and scatterer properties, ii) a new synthetic dataset for training based on the characteristics of real UK velocity maps, and iii) enhanced over-wrapping techniques with offset and gain to minimise the influence of reference point selection and to increase the likelihood of detecting slow deformation.
 
%%%%%%%%%%%%%%%%%%%%%%%%%%%%%%%%%%%%%%%%%%
\section{Materials and Methods}

\subsection{Convolutional Neural Networks}
\label{ssec:cnn}

Convolutional neural networks (CNNs) are a class of deep feed-forward artificial neural networks. They comprise a series of convolutional layers that are designed to take advantage of 2D structures, such as an image. The weights of the filter in each convolutional layer  are adjusted during the training process. The low-level features are extracted and connected to more semantic meaning at the deeper layers. In this paper, we want to learn features from the velocity maps that can distinguish deformation from stable ground.

Previous studies have used convolutional neural networks (CNNs) to detect deformation in wrapped InSAR images of volcanic environments \cite{Anantrasirichai:Application:2018,Anantrasirichai:deep:2019, Anantrasirichai:application:2019}. Wrapped interferograms are used because the high-frequency content of the fringes is easy to identify and provides strong features for the CNN. The work in \cite{Anantrasirichai:Application:2018} provided a proof of concept using a test dataset of 30,249 interferograms, compared different pre-trained networks and found AlexNet \cite{Krizhevsky:ImageNet:2012} to be the most effective and used data augmentation to train the network. The subsequent work \cite{Anantrasirichai:deep:2019} improved the detection performance by overcoming the lack of positive training data by using synthetic examples, representing deformation, turbulent and stratified atmospheric contributions. Recently, we studied the feasibility of using the CNN to detect slow volcanic deformation by rewrapping cumulative time series \cite{Anantrasirichai:application:2019}. We found that applying a gain of 2 to the interferograms to double the number of fringes can lower the detection threshold by 25–30\%, which can be as low as 1.3 cm/year.

 In this paper, we use a transfer-learning strategy augmented with fine-tuning the model trained in \cite{Anantrasirichai:deep:2019}.  Then the CNN model is retrained with some negative samples of the real UK data along with synthetic positive and negative samples, based on the characteristics of the real UK data as described in Section \ref{sec:Syntheticexamples}. In the prediction process, the velocity maps are wrapped and  converted into a grayscale image (i.e. the pixel values are scaled to $[0,255]$). Then they are divided into overlapping patches at the required input size for AlexNet (224$\times$224 pixels). Each patch is then repeatedly shifted (by 28=224/8 pixels in this paper) to cover the entire image.  The output of the prediction process is a probability $P$ of there being deformation in each patch. The probabilities from overlapping patches are merged using a rotationally symmetric Gaussian lowpass filter with a size of 20 pixels and standard deviation of 5 pixels.  

\subsection{UK InSAR dataset}
\label{subsec:ukinsar}

Fundamentally, all InSAR methods use the phase difference between two radar images to estimate changes in path length between the satellite and the ground surface. However, there are two distinct classes of processing approaches for generating time series of data: small baseline and persistent scatterer (PS). The small baseline technique \cite{Berardino:new:2002,Schmidt:time:2003} employs many small distributed scatterers and is commonly used for wide area monitoring, including tectonic and volcanic applications (e.g \url{http://comet.nerc.ac.uk/COMET-LiCS-portal/}). It produces a series of 2D images that can be straightforwardly employed by a CNN as shown by \cite{Anantrasirichai:application:2019}. In contrast, permanent or persistent scatterer methods \cite{Hooper:new:2004, Crosetto:Persistent:2016} focus on pixels dominated by a stable large reflector. Thus PS methods are well-suited to areas that have strong reflectors, especially man-made objects like buildings and are usually preferred for urban areas \cite{Lauknes:Comparison:2005}. However, the output dataset is sparse and not suitable for input into CNNs, where  correlations between adjacent pixels are learnt and used as local features for classification.

The InSAR dataset used in this paper was provided by SatSense Ltd who employ a novel pixel selection method, RapidSAR \cite{Spaans:InSAR:2016}. This technique works by identifying siblings of the selected pixel, i.e. evaluating nearby pixels with similar phase and amplitude to the selected pixel. This is then used to estimate the coherence of the selected pixel. This avoids the common issue with both persistent scatterer and small baseline methods whereby coherent points may be rejected or incoherent points included, due to the effect of surrounding pixels. The associated information loss and lower SNR is therefore avoided. However, this still corresponds to  sparse representation, which is not  directly suitable for CNNs. 

For the UK-wide study, we use the medium resolution SatSense product (10~m/pixel) for the period of 2015 - 2019 which consists of 66,801$\times$121,501 pixels. Although time series are available for each point, for this initial proof of concept, we simply use the average velocity for each pixel. In total, there are $\sim$64 million velocity measurements on the ascending pass and $\sim$29 million on the descending pass. The distribution of measurement locations is uneven with a significantly higher density in urban areas. We also identify three case study areas from the high resolution SatSense product (5~m/pixel). The coal mining area of Normanton and Castleford shows subsidence of more than 2 mm/yr (Fig \ref{fig:results}a) and South Derbyshire shows uplift of more than 6 mm/yr (Fig \ref{fig:results}d). A linear pattern of subsidence is seen from Battersea Power Station to Kennington in London (Fig \ref{fig:results}g). This is the Northern line extension, where two 3.2~km tunnels have been created between 2017-2020. The difference between the two resolutions is illustrated in Fig. S1 in the supplementary material.

\begin{figure}[t!]
	\centering
      	  \includegraphics[width=\columnwidth]{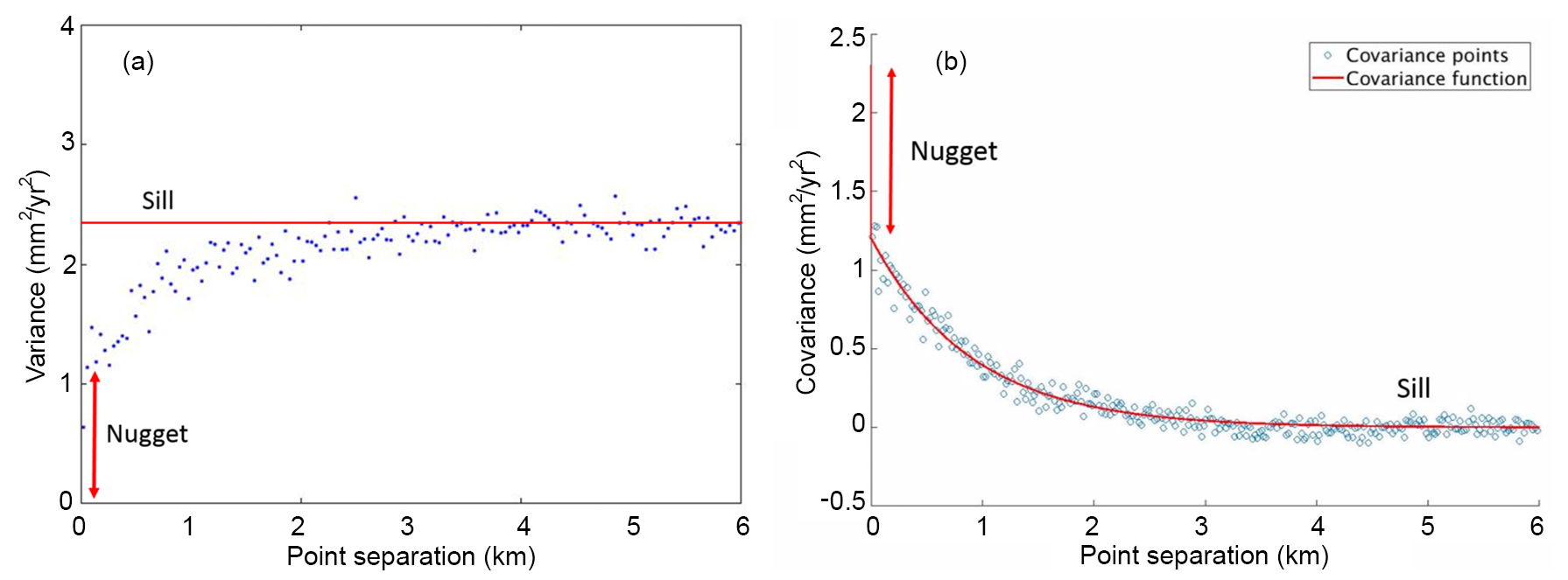}  
					\caption{Spatial characteristics of data in Easton, UK. (a) A 6~km length variogram, showing the nugget near 0, then levels off to a sill of 2.3 mm$^2$/yr$^2$ at point separations of above around 2 km.  (b) A covariance function, showing the exponential fitted to the data, the nugget at zero point separation, and the sill as the function levels off at point separations greater than $~$2 km.}
    \label{fig:S2_variogram}
\end{figure}

To analyse the spatial characteristics of the SatSense dataset in the UK, we performed a spatial analysis using covariogram \cite{Wackernagel:Multivariate:2003}. First, a spatial variogram $\gamma(d)$ for point velocity values in space is computed, where $d$ is the distance between the pixels. We found that the variance of point velocity increases sharply (the nugget $\mu_{nugget}$) when the distance between the points is close to zero, then exponentially increases and exhibits a sill $\mu_{fill}$, the background variance value, at long length scales. Consequently, a theoretical variogram is related to a covariance $C(d)$ on the basis of $\gamma(d) = \mu_{sill} -  C(d)$. That is, the covariance $C(d)$ decreases exponentially, which is expressed as 
\begin{equation}
\label{eq:cov}
	C(d) = 
	\begin{cases}
	a e^{-bd}, & \text{if}  \: d > 0 \\
	\mu_{sill} , & \text{if}  \: d=0
	\end{cases}
\end{equation}
\noindent where $a$ and $b$ are constants, $a=\mu_{sill} - \mu_{nugget}$, and $d$ is the separation distance in km. From the available UK dataset, we found  $a = 0.7- 1.8$~mm$^2$/yr$^2$,  $b  = 0.8-1.6$, and $\mu_{sill}=1.5-2.9~$mm$^2$/yr$^2$. This appears as spike noise in the InSAR image and disturbs the gradient calculations performed by the CNN. Thus the spike noise needs to be accounted for when addressing the issue of data sparsity. The plots of the variogram and covariance are shown in Fig. \ref{fig:S2_variogram}.

% ====================================================
\section{Theoretical Contributions}
\label{ssec:solutions}

%{\bf I've removed a paragraph here for now, but might reinstate it later. It seemed we'd mentioned this already in the introduction, so ti was a bit repetetive}
% CNNs are data-driven approaches and are perhaps the most powerful, machine learning tools for image classification and recognition. However, as stated above the UK InSAR data is sparse and analysis of the spatial characteristics shows a 'nugget' in the variogram at zero point separation. Thus the characteristics of the UK datasets are very different to the interferograms used for previous applications of pre-trained CNNs.   The UK sources of deformation are anthropogenic and can attain thicknesses of $>$65~m in the UK, but component units are commonly only a few metres thick \cite{Ford:assessment:2014}. This means the deformation patterns are significantly more localised than those in volcanic examples. Moreover, the successful CNN require a large training set of labelled data. We therefore propose three enhanced solutions (described below) to overcome these requirements.

% --------------------------------------------------------------------
\subsection{Spatial interpolation}
\label{sec:Spatialinterpolation}

\begin{figure*}[t!]
	\centering
      		 \includegraphics[width=0.8\textwidth]{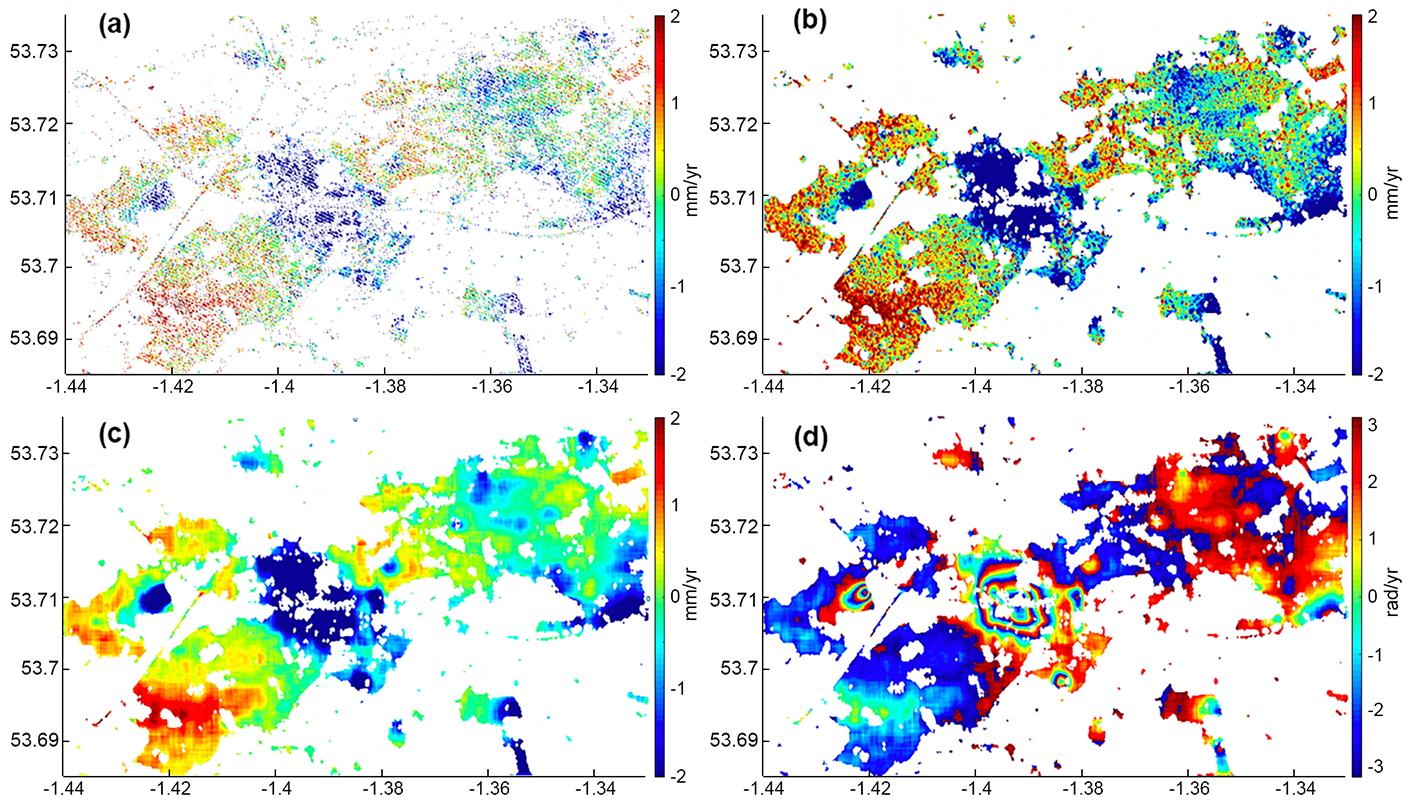}  
					\caption{\small Velocity map at Normanton and Castleford showing (a) raw sparse data, and its interpolated results from (b) Delauney Triangulation and (c) Matrix Completion techniques. The wrapped velocity map of (c) with the wrap gain $\mu$=8 is shown in (d). }
    \label{fig:matrixcompletion}
\end{figure*}

CNNs rely on spatial or sequential attributes of dense data to learn effectively. Adjacent pixels share information that is important and the inherent structure to pixels in image data gives meaning to the overall image. If the data is highly sparse, then the network learns `zeros', the gradient of the loss function is zero and the performance does not improve with iteration. Therefore, it is necessary to interpolate the data during pre-processing to resemble a dense image. Here, we propose and test a novel interpolation method specifically for sparse InSAR data. We illustrate the process using the case study of Normanton and Castleford as shown in Fig. \ref{fig:matrixcompletion} a-c and test the ability of the CNN to identify signals for different types of interpolation in section \ref{ssec:spatialresult}.

The simplest way to mathematically describe sparse images is by $y = Mx + n$, where $y$ is the sparse observation of an ideal dense signal $x$, $M$ is the sub-sampling matrix, which can be seen as a mask of existing or non existing values, $n$ is noise. Here $y$ is the raw velocity measurements shown in Fig. \ref{fig:matrixcompletion}a). This  poses an inverse problem for finding $x$. We employ a matrix completion method (MC) which has been used for compressive sensing \cite{Yang:SAR:2014}, where the sparsity of a signal can be exploited to recover it from far fewer samples than required by the Nyquist Shannon sampling theorem  \cite{Candes:matrix:2010}.  
This can be solved with an optimisation process as 
\begin{equation}
\label{eq:optimisation}
	\hat{x} = \arg \min_x \{\frac{1}{2}||y-M x||_2^2+\alpha ||x||_*\},
\end{equation}
\noindent where  $||x||_*$ is nuclear norm of a matrix (a convex hull of the rank function of $x$) and $\alpha$ is a regularization parameter. This can be done through a non-convex  matrix completion via iterated soft thresholding \cite{Cai:svt:2016}. The nuclear norm is computed using singular values of matrix $x$ and the process tries to achieve 
\begin{equation}
\label{eq:svt}
	\min_x ||S_x||_p \: \: \text{subject to} \: \: ||y - M x||_2<\varepsilon,
\end{equation}
\noindent where  $U_x,S_x, V_x = \text{SVD} (x)$,  $\text{SVD}$ is singular value decomposition giving the outputs such that $x= U_x S_x V'_x$ and for a non convex function, $0<p<1$. The pseudocode to describe this optimisation process is given in Algorithm \ref{al:Pseudocode}.

First, we generate an initial $x_{0}$ by first suppressing some high noise and then applying Delaunay triangulation (DT) (Fig. \ref{fig:matrixcompletion}b). To suppress the high-amplitude noise, we simply apply a two-dimensional median filter $\text{Med}_{3\times 3}(\bullet)$ that omits NaN values in the median calculation. We record the noise map $N = y - \text{Med}_{3\times 3}(y)$, which will be used later for generating synthetic data with similar characteristics (Section \ref{sec:Syntheticexamples}).

In the interpolation process, we add a Gaussian filter $G(x,\sigma)$ with standard deviation $\sigma$ of 5 pixels, to remove the remaining spike noise in each iteration loop. The proposed technique achieves the estimation of missing pixels and noise reduction simultaneously. Figure \ref{fig:syntheticexample} shows that the proposed matrix completion method produces more realistic results than conventional Delauney triangulation alone.

 \begin{algorithm}[t!]
 \caption{Pseudocode of optimization algorithm}
 \label{al:Pseudocode}
 \small
 \begin{algorithmic}[1]
 \renewcommand{\algorithmicrequire}{\textbf{Input:}}
 \renewcommand{\algorithmicensure}{\textbf{Output:}}
 \REQUIRE $y$, $x_{0}$, $f_{0}$, $p$, $\alpha_0$, $\alpha$, $\lambda$, $\tau $, $K$\\
 $y:$  sparse observation \\
$x_{0}:$ interpolation using DT and noise suppression \\ 
$M:$ sub-sampling matrix \\
$\alpha:$  regularization parameter, $\alpha_0 = 0.9 \max(|M x|)$ \\
$f$: loss, initialled with $f_{0}=|| y - M x_0||_2 + \alpha_0  || x_0 || $ \\
$p:$  non-convex norm,  $p=$0.8 \\
$\lambda:$ $1.1\cdot$eigenvalue of $(M^{-1} M)$\\
$\tau:$ tolerance, $\tau=10^{-4}$ \\
$C$: cost function \\
$K:$ maximum iterations, $K=$200
 \ENSURE  $\hat{x}=x$
  \WHILE {$\alpha > \tau \alpha_{0}$}
  \FOR {$k = 1$ to $K$}
  \STATE $x \gets x + \frac{1}{\lambda}M^{-1}(y - M x)$
  \STATE $U$, $S$, $V \gets$ SVD$(x)$
  \STATE $S \gets$ diag$\{S\}$
  \STATE $S \gets$ sign$(S) \max(0,|S|-\frac{1}{2\lambda}\alpha |S|^{p-1}$)
  \STATE $x  \gets U($diag$\{S\})V'$
  \STATE $x \gets G(x, \sigma)$ 
 \STATE  $f_k \gets || y - M x||_2 + \alpha  || x || $
 \STATE  $C \gets {|| f_k - f_{k-1} || } / { || f_k + f_{k-1} ||} $
 \IF {$C < \tau$}
 \STATE break
 \ENDIF
 \ENDFOR
 \STATE $\alpha \gets 0.9 \: \alpha$
  \ENDWHILE
 \end{algorithmic} 
 \end{algorithm}
 
% --------------------------------------------------------------------
\subsection{Synthetic examples}
\label{sec:Syntheticexamples}

\begin{figure}[t!]
	\centering
     		\includegraphics[width=\columnwidth]{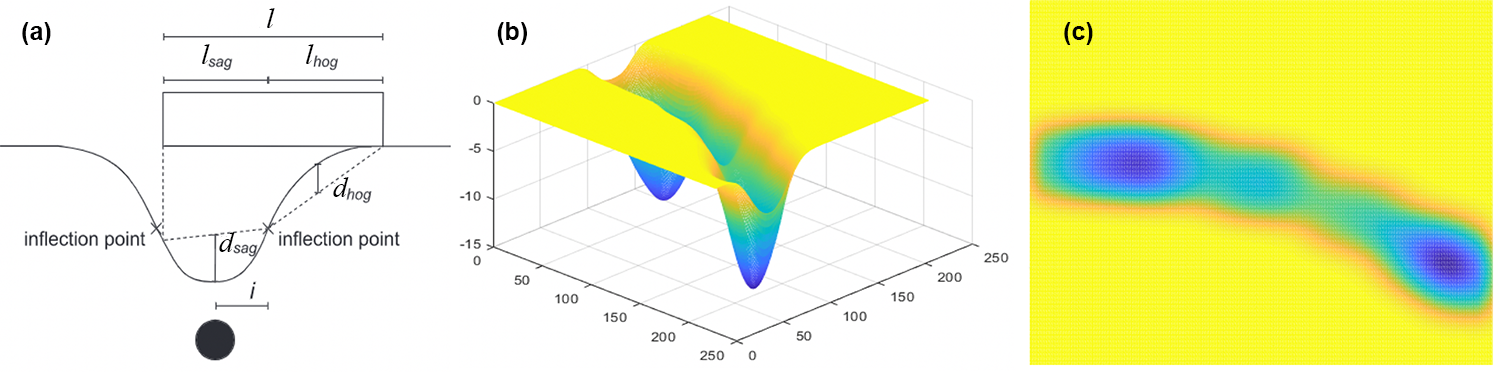}  
					\caption{Synthetic tunnelling subsidence generated following the model introduced in \cite{Giardina:Evaluation:2018}, where the cross section profile is shown in (a). Our three-dimensional (3D) synthetic deformation and its projection to create two-dimensional (2D) unwrapped velocity map are shown in (b) and (c), respectively.}
    \label{fig:syntunnelling}
    \vspace{2mm}
    \includegraphics[width=\columnwidth]{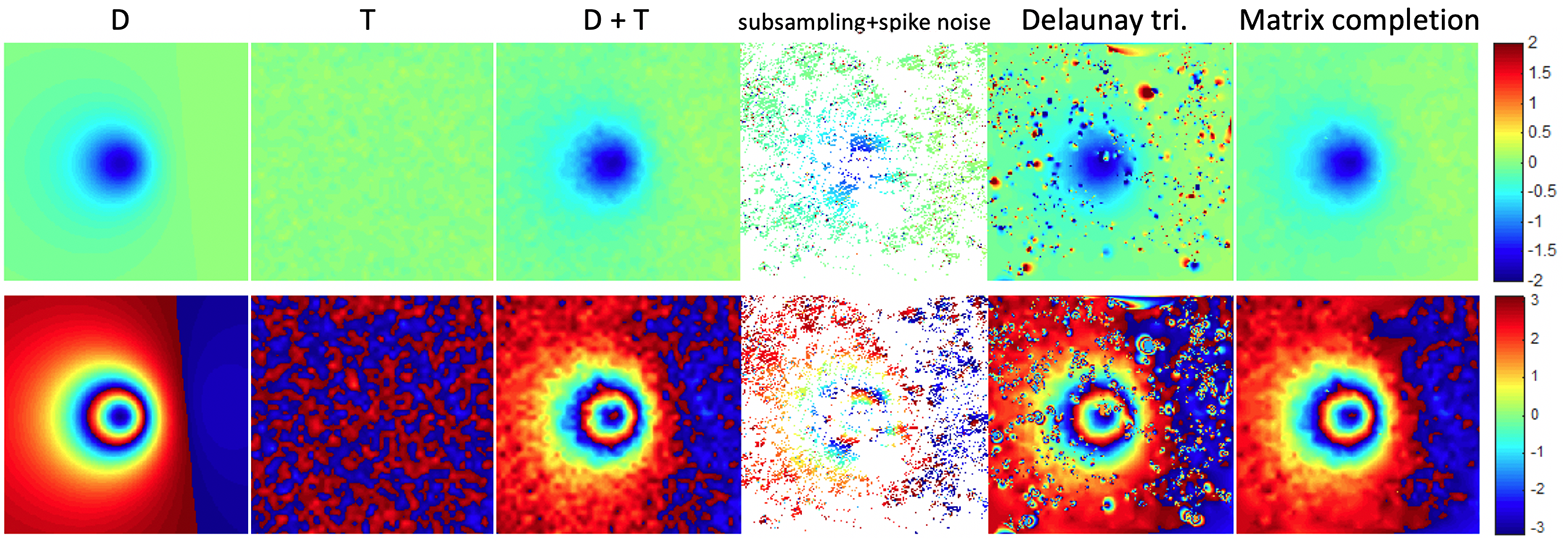}  
					\caption{\small Synthetic example showing (top row) unwrapped  and (bottom  row) wrapped samples.}
    \label{fig:syntheticexample}
\end{figure}

We create 10,000 synthetic datasets ($X$) for training the CNN using 2 components, namely deformation $D$, and turbulent atmosphere $T$, using the simple linear function $X = D+T$. Figure \ref{fig:syntheticexample} demonstrates the process of synthetic example generation for one example.  In this paper, we concentrate on deformation caused by coal mining and tunnelling as they are common in the UK. Therefore we employ two models as follows.
	i) \textit{A set of synthetic examples of coal mining subsidence}: $D_{point}$, is generated using a point pressure source model \cite{Mogi:Relation:1958}, which reproduces the surface deformation associated with inflation and deflation of a subsurface point source. To represent the shallow sources associated with coal mining, we use depths of 3 - 80~m and volume changes of $10^{0.3} - 10^3$ m$^3$.
	ii) \textit{A set of synthetic examples of tunnelling subsidence, $D_{line}$} is generated following \cite{Giardina:Evaluation:2018}. The tunnelling-induced subsidence profile is modelled with sagging and hogging zones as demonstrated in Fig. \ref{fig:syntunnelling}a, where the length and depth parameters of sagging and hogging zones are $l_{sag}$, $l_{hog}$, $d_{sag}$ and $d_{hog}$, respectively. 
We use both $l_{sag}$ and $l_{hog}$ in a range of 30 - 80~m, $d_{sag}$ of 1 - 10~mm, and $d_{hog}$ of 1 - 5~mm. $D_{line}$ is generated by varying these parameters along the curve and straight lines, replicating the track of the underground rail. The 3D displacement vector is then projected to line of sight (LOS) using Sentinel-1 UK incidence and heading angles for ascending and descending passes.
For both cases, the range of parameters is chosen so that the LOS velocity is in the range 0-15 mm/yr. Note that, in this paper, we trained the models of $D_{point}$ and $D_{line}$ separately, but they could be merged to train a 3-class model (2 types of deformation and non-deformation) in the future.

The satellite measurements of displacement are affected by atmospheric delays caused primarily by water vapour in the troposphere, $T$. The delays are spatially correlated and their covariance is described in Section \ref{subsec:ukinsar}. For simplicity, the statistical properties of the atmosphere are assumed to be radially symmetric and have a homogeneous structure \cite{Parsons:1994:2006}. We use Monte Carlo samples of these distributions to generate synthetic variance-covariance matrices and use a Cholesky decomposition to produce synthetic images with the corresponding statistical properties \cite{Anantrasirichai:deep:2019}. For previous applications to volcanic environments, we have also considered a stratified atmospheric component related to the high relief of volcanic edifices. This effect is small in the UK and is neglected here.

We then sub-sample the combined image ($D$+$T$) using randomly chosen distributions of points from the SatSense data and add spike noise as described in Section \ref{subsec:ukinsar}.  Finally the sparse signals are interpolated as described in Section \ref{sec:Spatialinterpolation}.

% --------------------------------------------------------------------
\subsection{Overwrapping and phase shifting }
\label{ssec:Overwrapping}

 We wrap the velocity map to provide strong features for machine learning \cite{Anantrasirichai:Application:2018}. To deal with different deformation rates, we  combine a range of wrapping intervals following the method of \cite{Anantrasirichai:application:2019}, which was originally designed to detect slow, sustained volcanic deformation in time-series data, but can be adapted for detecting slow, localised motion in the UK velocity measurements.  Theoretically,  the number of fringes can be increased without altering the signal to noise ratio by reducing the wrap interval ($\mu$). In this paper, following Sentinel-1 line-of-sight where one fringe represents 28 mm of displacement, we employ wrap intervals of 14~mm/yr, 7~mm/yr, 3.5~mm/yr, and 1.75~mm/yr in the prediction process.
 
One problem with wrapping the velocity map is that different reference points cause the wrap discontinuities to occur in physically arbitrary locations.  For some choices of reference points, the number of fringes will increase, but for others it will decrease or for very small signals, fail to produce any discontinuities at all. To ensure that fringes exist on the test image, a constant offset $\tau$ is added  to the velocity map $\psi$ producing $\psi'_\tau$, i.e. $\psi'_\tau  \equiv   \psi + \tau  \mod{\mu}$. We run 4  offsets, and select the maximum probability from the CNN for each wrap interval $\mu$, i.e. $P_{\mu}$ =  $\max \{ P_{\mu,\tau} \}, \: \: \tau \in \{0, 3.5, 7, 10.5 \}$~mm/yr, and $\mu \in \{14, 7, 3.5, 1.75 \}$~mm/yr. The final result is the average of the four probabilities, i.e. $P_{\text{final}} = \frac{1}{4} \sum_{\forall \mu} P_\mu$.

%Following Sentinel-1 line-of-sight, one fringe represents 28 mm of displacement, so we convert the velocity $v$ in the phase $\psi$ by $\psi = 2 \pi v /28 $. We also employ shifting the wrapping boundaries by adding a constant phase offset $\tau$ to the velocity map using $\psi'_\tau$, i.e. $\psi'_\tau  \equiv   \psi + \tau  \mod{2\pi}$. This reduces the offset due to different reference points. The phase discontinuities however occur in physically arbitrary locations, so for some cases, the number of fringes will increase, but in others it will decrease. Therefore, we run 4 phase offsets, and select the maximum probability from the CNN for each gain $\mu$, i.e. $P_{\mu}$ =  $\max \{ P_{\mu,\tau} \}, \: \: \tau \in \{0, \pi/2, \pi, 3\pi/2 \}$. We run the prediction for $\mu=2^m, m \in \{1,2,3\}$ and the final result is the average of these four probabilities.

% {\bf because we're starting from the velocity measurements, they were never 'wrapped' in the first place, and it's not obvious what the original wrap interval should be. Would be better to write this in terms of a range of wrap intervaferls e.g. 1mm/yr, 2mm/yr, 4mm/yr 8 mm/yr. Not sure what you actually used though.}

% --------------------------------------------------------------------
\subsection{Combining different line of sight geometries}
\label{ssec:combineLOS}

One limitation of InSAR technology is that the ground motions are measured in a one-dimensional line of sight (LOS) geometry, whilst the actual surface motions can occur in three dimensions. This means the deformation detected in one LOS direction might not be able to be detected in another LOS direction. However, an advantage is that noise causing a false positive result that appears in one acquisition might not affect the acquisition in another LOS. Therefore in this study, if the areas have both ascending and descending passes available, the two velocity maps are processed independently and the final probability results are obtained from the average. If there are four looks (2 ascending and 2 descending passes), the final probability map will be the maximum of four averages between a pair of ascending and descending signals.

%%%%%%%%%%%%%%%%%%%%%%%%%%%%%%%%%%%%%%%%%%
\section{Results and Discussion}
\label{sec:Results}

\subsection{Spatial interpolation}
\label{ssec:spatialresult}

We first investigate the performance of the proposed spatial interpolation technique using synthetic datasets. Three approaches are tested i) sparse examples without interpolation, ii) interpolated examples with Delauney Triangulation (DT), and iii) interpolated examples using the proposed Matrix Completion (MC) approach (see Fig. \ref{fig:syntheticexample} last three columns). The CNNs are trained with two classes: $D+T$ (positive) and $T$ (negative). Each class contains 10,000 synthetic samples.  When training the CNN with sparse examples, the results of convolution processes are computed from the pixels that have values only. The classification results are shown in Table \ref{tab:results_syn}. It is obvious that without spatial interpolation, the CNN cannot distinguish between deformation and non-deformation (the accuracy is around 50\%). The CNN performs significantly better with dense datasets with an improvement of accuracy by 64.0\% with the initial DT and 81.5\% with the proposed MC. The DT produces 10 times more false positives  than the MC due to spike noise (the nugget - see Section \ref{subsec:ukinsar}).

\begin{table}
	\centering
	\caption{Classification performances (\%) when training with sparse and interpolated examples}
	\small
		\begin{tabular}{ccccc}
		\toprule
        Dataset &  Accuracy &  Precision & Recall & False positive rate\\
	    \midrule
        	Sparse & 54.32 & 63.91 & 53.62 & 55.27 \\
        	Interp. DT & 89.06 & 99.10 & 82.52 & 20.98\\
        	Interp. MC & 98.58 & 99.27 &  97.93 & 2.09\\
        \bottomrule
		\end{tabular}
	\label{tab:results_syn}
\end{table}

\subsection{Application to case study sites}
\label{ssec:regionresult}

\begin{figure*}[t!]
	\centering
      	  \includegraphics[width=\textwidth]{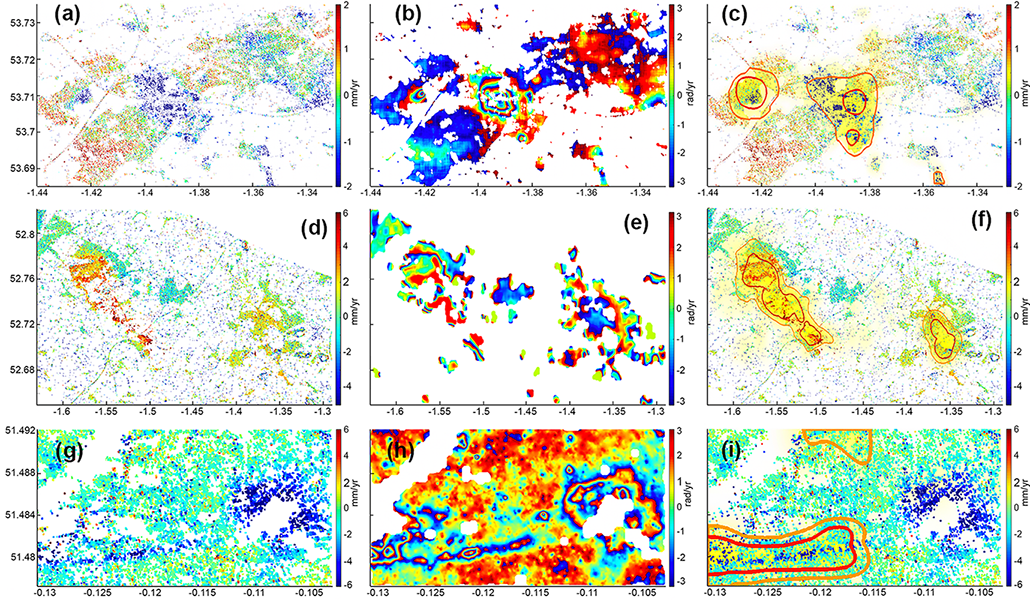}  
					\caption{\small Detection results in (a-c)  Normanton and Castleford, (d-e) South Derbyshire, and (g-i) London -- Northern line extension. (a), (d) and (g) are raw data. (b), (e) and (h) are the wrapped and interpolated velocity maps. (c), (f) and (i) are probability maps overlaid on the raw data. The brighter yellow means higher probability. Areas inside orange and red contours are where $P>$0.5 and $P>$0.75, respectively. }
    \label{fig:results}
\end{figure*}

Initially, we test our machine learning algorithms on well-known case study examples of coalfield subsidence $D_{point}$ and tunnelling $D_{line}$ (as described in Section \ref{sec:Syntheticexamples}) using the high resolution InSAR product (5~m/pixel). The models are trained separately, using the synthetic examples. 
The  detection results are shown in Fig. \ref{fig:results}, where 
the first, the second and the third columns show i) raw InSAR data, ii) wrapped and interpolated velocity maps used as inputs of the CNNs, and iii) the probability values overlaid on the velocity maps, respectively. The first and the second rows are the results from the coalfields at Normanton and Castleford, and South Derbyshire, detected with the model $D_{point}$. The velocity map at South Derbyshire has fewer data points causing more difficulties for the interpolation step than that at Normanton and Castleford, but the detection algorithm still works well in both cases.
The last row of Fig. \ref{fig:results} shows the detected tunnelling subsidence in London using the model $D_{line}$. Interestingly the model detects the line of the tunnel but do not pick out the point-source deformation (on the right of the image). These case study results are promising and warrant further testing to check the generalisation of the model and the applicability to a larger scale map.

\subsection{Whole UK velocity map}
\label{ssec:wholeUK}
\begin{figure*}[t!]
	\centering
      	  \includegraphics[width=\textwidth]{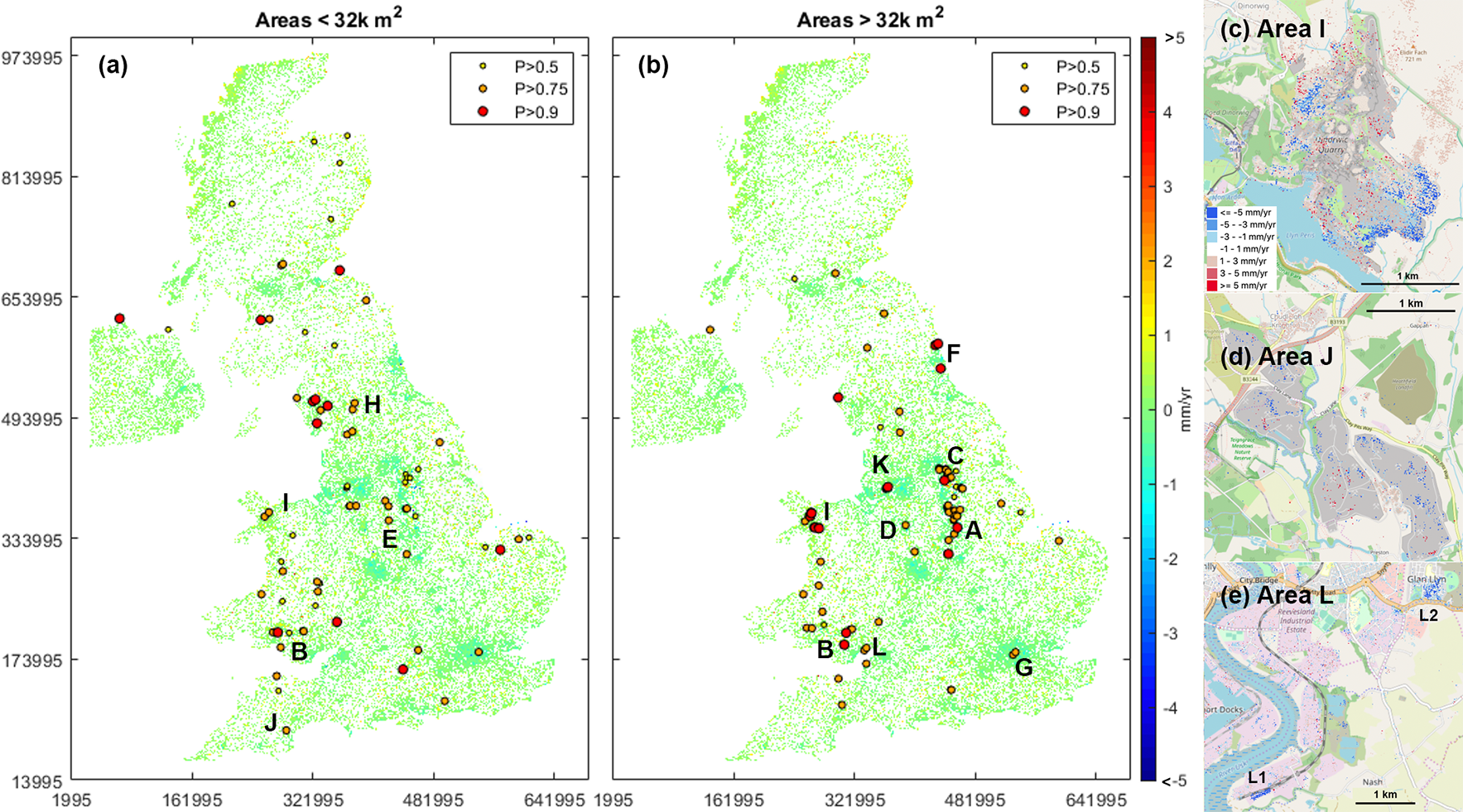}  
					\caption{\small Detection results in the 10-m velocity map showing the centre locations where the CNN using $D_{point}$  identifies with high probability of being deformation. For visualisation, the small and large areas are plotted separately: (a) the area size less than 32~km$^2$, and (b) the area size larger than 32~km$^2$. Right column shows  ground subsidence due to anthropogenic sources at (c) the Dinorwic quarry in North Wales (Area I), (d) the clay works in Kingsteignton (Area J), (e) the coal yard of Uskmouth power station (L1) and residential areas around Brinell Square (L2)  in Newport (Area L).}
    \label{fig:UK_map_results}
\end{figure*}

\begin{figure*}[t!]
    \includegraphics[width=0.875\textwidth]{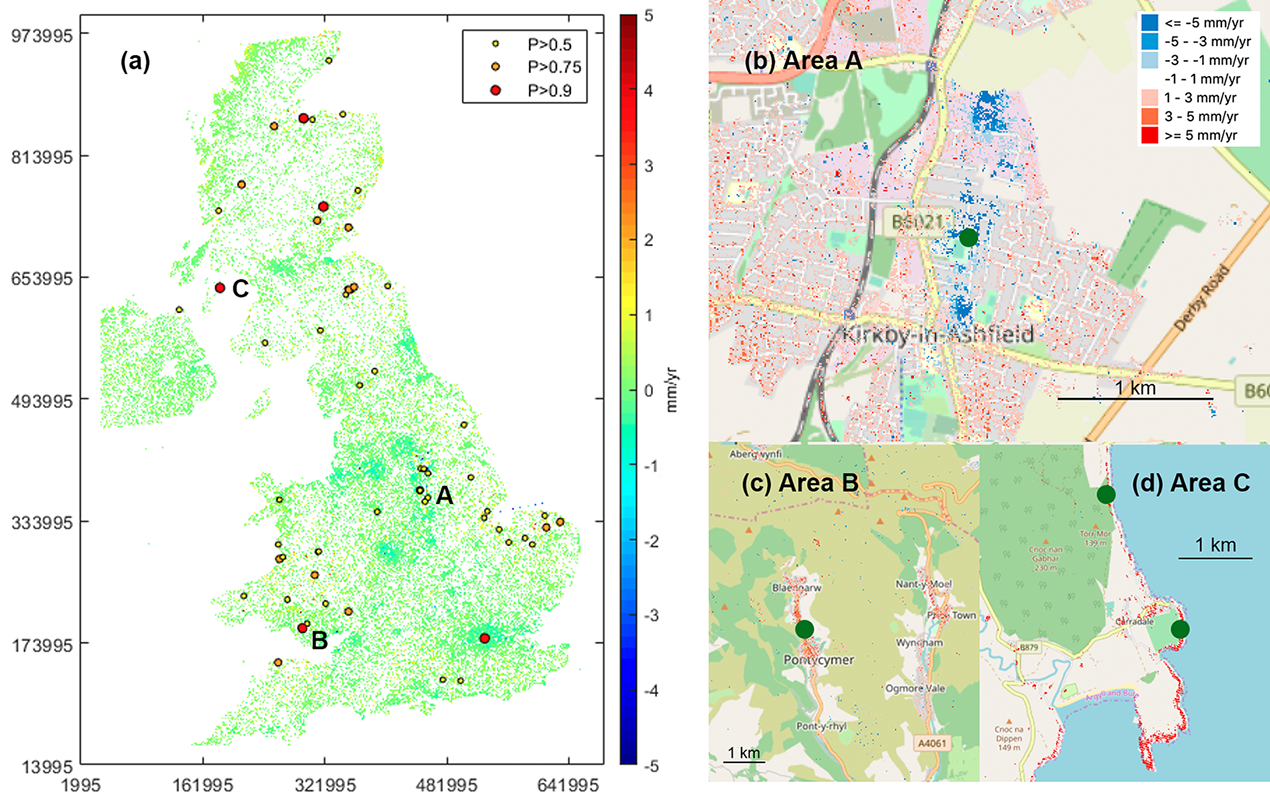}  
					\caption{\small  Detection results in the 10-m velocity map (a)  showing the centre locations where the CNN using $D_{line}$  identifies with high probability of being deformation. Examples of linear deformation in (b) Kirkby-in-Ashfield (Area A), (c) Pontycymer, Wales (Area B),  (d) uplift coastline in Carradale, Scotland (Area C). The green dots on (b)-(d) are the centre of detected areas.}
    \label{fig:UK_map_results_tunnels}
\end{figure*}

As described in Section \ref{subsec:ukinsar}, there are $\sim$64 million points of sparse UK data. This is equivalent to a 2D image with a resolution of 98,504$\times$68,504 pixels, which is more than 3,250 full HD TVs combined.
To automatically process this large velocity map, we divide it into several 2500$\times$2500  maps, defined by the limitation of memory required to process the spatial interpolation. After spatially interpolating,  each velocity map is further divided into overlapping patches following the detection process described in Section \ref{ssec:cnn}.
The detected deforming locations using model $D_{point}$ and $D_{line}$ are plotted in Fig. \ref{fig:UK_map_results} and Fig. \ref{fig:UK_map_results_tunnels}, showing three levels of probability $P$, which are $>$0.5, $>$0.75 and $>$0.9. In the supplementary material (Fig. S2 and S3) we show areas with detection probabilities $>$0.5 in more detail.

Fig. \ref{fig:UK_map_results} shows the results of the $D_{point}$ model. 
The method detects numerous deforming areas in well-known coal-mining regions from the Midlands up towards Leeds (area A in Fig \ref{fig:UK_map_results}), in South Wales (area B \cite{Bateson:application:2015}), Normanton and Castleford (area C), North Staffordshire in Stoke-on-Trent (area D \cite{Culshaw:Measurement:2006}), Northwest Leicester (area E \cite{Sowter:DInSAR:2013}), Northumberland and Durham (area F \cite{Gee:Ground:2017}). Several areas are detected in London, where recent engineering work has taken place. For example, the detected uplift at Canning Town, London, could be affected by groundwater rebound after completion of dewatering works for the underground construction (area G \cite{Boni:Methodology:2018}).  In the northwest of Wales, the method detects subsidence from some former slate quarries (area I), including the Dinorwic Quarry near Llanberis (Fig. \ref{fig:UK_map_results}c), the Penrhyn Quarry near Bethesda, and the Ffestiniog Slate Quarry in Blaenau Ffestiniog, where the slate was mined rather than quarried. The method also detects subsidence of clay works in Kingsteignton (area J, Fig. \ref{fig:UK_map_results}d).
Uplift was detected at Golborne, Leigh and Manchester (area K) with a similar spatial extent to the subsidence reported between 1992–2000 \cite{Cigna:relationship:2017}. 

 Although we are dominantly considering vertical deformation, horizontal motion associated with landslides and coastal processes will also cause displacements in the line of sight (see Fig. S4 in the supplementary material). For example, landslides with significant horizontal motion were detected south of Kirkby Stephen (area H \cite{Novellino:Assessing:2017}).

Fig. \ref{fig:UK_map_results_tunnels} shows the results of the $D_{line}$ model. We did not include examples of uplift in either positive or negative training datasets for $D_{line}$, but nonetheless, we detect several uplifting features because the fringes in the wrapped velocity map have characteristics closer to the positive samples than the negative ones. Since uplift and subsidence can be simply distinguished by comparing the velocity with that of neighbouring areas, this information can be added in post-processing. The only detection of tunnelling subsidence in London was at the case study site shown in Fig. \ref{fig:results}g-\ref{fig:results}i, but there were several detections elsewhere in the UK, particularly in the Midlands. Several of these are elongated areas of subsidence more in keeping with mining (for example following coal seams) than infrastructure tunnels (Area A in Fig. \ref{fig:UK_map_results_tunnels}).  In several cases, linear features are associated with linear surface structures, such as roads, probably due to the higher density of measurement points on the man-made structures than in the surrounding fields. The deformation signal itself is unlikely to be linear, but this enables us to identify deformation sources that might be missed by the $d_{point}$ model due to the uneven sampling of data (e.g.  Fig. \ref{fig:UK_map_results_tunnels}c). In several places, rocky foreshores are picked out, such as the coastline in Carradale (Fig. \ref{fig:UK_map_results_tunnels}d), which appears to be uplifting relative to the nearest inland point. We attribute this to processing artefacts within the InSAR data.

%%%%%%%%%%%%%%%%%%%%%%%%%%%%%%%%%%%%%%%%%%
\subsection{Discussion}

Monitoring ground deformation is crucial in urban and semi-urban areas. The UK has a long history of coal mining, and associated water pumping causes surface deformation which can extend to city-sized regional areas. Slope instability can lead to localised damage both in hilly areas and coastal regions. Ground motion can have negative impacts on infrastructure, particularly long linear assets such as  drainage networks and pipelines.  
An example of the need of ground movement detection is for the proposed HS2 route for high speed rail\footnote{https://www.hs2.org.uk} from Birmingham to Leeds, which would pass through the large coalfield areas in Nottingham and Sheffield. Fig. \ref{fig:UK_map_results} and Fig. \ref{fig:UK_map_results_tunnels} show clear ground deformation in these areas and although the velocity rate is only millimetres per year, it still needs to be factored into construction plans.

%Previous work on deep learning has demonstrated its potential for automatically searching through large volumes of wrapped InSAR images to detect both slow and rapid ground deformation that may be related to volcanic activity. In this paper, we extend our work to the UK velocity map, where the measurement points are sparse and unevenly distributed. Moreover, the spatial noise characteristics of this dataset are different from the distributed scatterer InSAR used in the volcano case. We analysed this type of InSAR data and propose several new adaptations to allow the transfer learning approach to perform well under these circumstances. 

This paper is a proof-of-concept that demonstrates the potential applicability of the deep learning framework to the development of automated ground motion analysis for anthropogenic sources of deformation in urban and semi-urban environments. We test the deep learning framework on the UK dataset and produce a probability map of surface movement. As the dataset is very large (see Section \ref{sec:Results}), it would not be feasible to manually inspect the entire area at high resolution.  Using a probability threshold of 0.5, the method produces some false positives and false negatives. However, the probability values and the sizes of the detected areas can be employed to prioritise further analysis.

This approach is not restricted to the UK dataset and could be used for any national or regional velocity map, including the European Ground Motion Service currently proposed by Copernicus \cite{MSUCSE062}. The main limitation of the current framework is that it cannot detect very localised deformation, like sinkholes, because their spatial characteristics are  too similar to noise.  These areas however show clear changes in the time domain. Future developments can incorporate both time-series analysis and spatio-temporal (3D) analysis of InSAR data. Moreover, if both ascending and descending passes are available for the same period of time, 4D signals can be used. In this paper, we train the model using only one pass (2D), and the results of both passes are averaged (Section \ref{ssec:combineLOS}). If both passes are concatenated and trained together, we expect that the deformation signals would be shown in both passes, so the number of false positives arising from using only one pass will be diminished.

%%%%%%%%%%%%%%%%%%%%%%%%%%%%%%%%%%%%%%%%%%
\section{Conclusions}

This paper demonstrates the feasibility of using a  transferable CNN approach to detect ground deformation in urban and semi-urban areas in the UK. We analyse characteristics of the data and propose several adaptations to previously developed deep learning methods.  Matrix completion is  used to overcome the sparse and uneven measurement distribution and simultaneously reduce spike noise. Synthetic examples based on point sources and tunnels are used for training due to lack of real signals of deformation. Finally  overwrapping and phase shifting techniques are employed to enhance features and hence reduce the detection threshold. The methods are tested using the velocity map generated by SatSense Ltd. dated between 2015-2019 and  successfully detect several types of deformation occurring around the UK.

\section*{Acknowledgment}
The authors would like to thank SatSense Ltd. for providing UK datasets.

%%%%%%%%%%%%%%%%%%%%%%%%%%%%%%%%%%%%%%%%%%
%% optional
%\appendixtitles{no} %Leave argument "no" if all appendix headings stay EMPTY (then no dot is printed after "Appendix A"). If the appendix sections contain a heading then change the argument to "yes".
%\appendix
%\section{}
%\unskip
%\subsection{}
%The appendix .

%\section{}
%All appendix sections must be cited in the main text. In the appendixes, Figures, Tables, etc. should be labeled starting with `A', e.g., Figure A1, Figure A2, etc. 


% The following MDPI journals use author-date citation: Arts, Econometrics, Economies, Genealogy, Humanities, IJFS, JRFM, Laws, Religions, Risks, Social Sciences. For those journals, please follow the formatting guidelines on http://www.mdpi.com/authors/references
% To cite two works by the same author: \citeauthor{ref-journal-1a} (\citeyear{ref-journal-1a}, \citeyear{ref-journal-1b}). This produces: Whittaker (1967, 1975)
% To cite two works by the same author with specific pages: \citeauthor{ref-journal-3a} (\citeyear{ref-journal-3a}, p. 328; \citeyear{ref-journal-3b}, p.475). This produces: Wong (1999, p. 328; 2000, p. 475)

%=====================================
% References, variant B: external bibliography
%=====================================
\bibliographystyle{IEEEtran}
\bibliography{machinelearning}


%\vfill


\newpage

\centering{
\begin{huge}
 \textbf{Supplementary Material \\ Deep Learning Framework for Detecting Ground Deformation in the Built Environment using Satellite InSAR data}
\end{huge}}

\vspace{5cm}

\renewcommand\thefigure{\thesection.\arabic{figure}}  
\renewcommand{\thesection}{S\arabic{section}}  
\renewcommand{\thetable}{S\arabic{table}}  
\renewcommand{\thefigure}{S\arabic{figure}}

\setcounter{figure}{0}

\begin{figure*}[h!]
	\centering
      	  \includegraphics[width=0.5\textwidth]{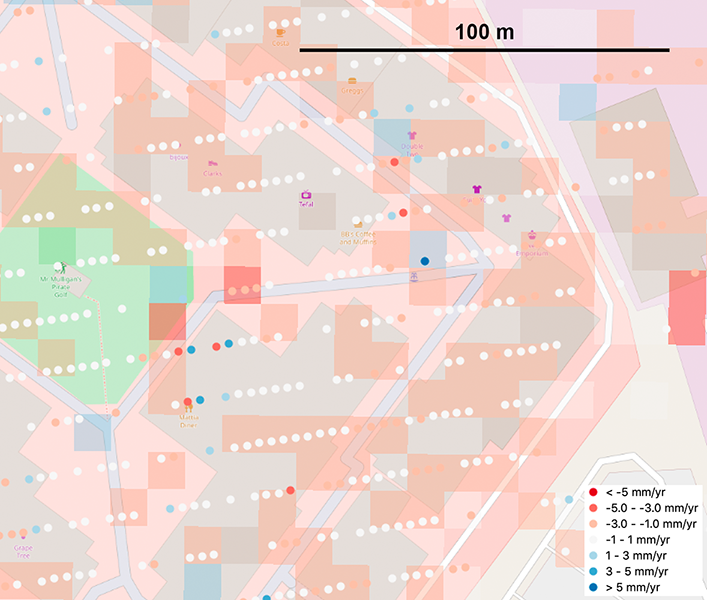}  
					\caption{Comparing the coverage of one pixel in medium- (10~m/pixel) and high-resolution (5~m/pixel) products from Satsense Ltd. The points of medium resolution are shown in square blocks, and the points of high resolution are shown in dots. }
    \label{fig:compare_resolutions}
\end{figure*} 

\begin{figure*}[h!]
	\centering
      	  \includegraphics[width=\textwidth]{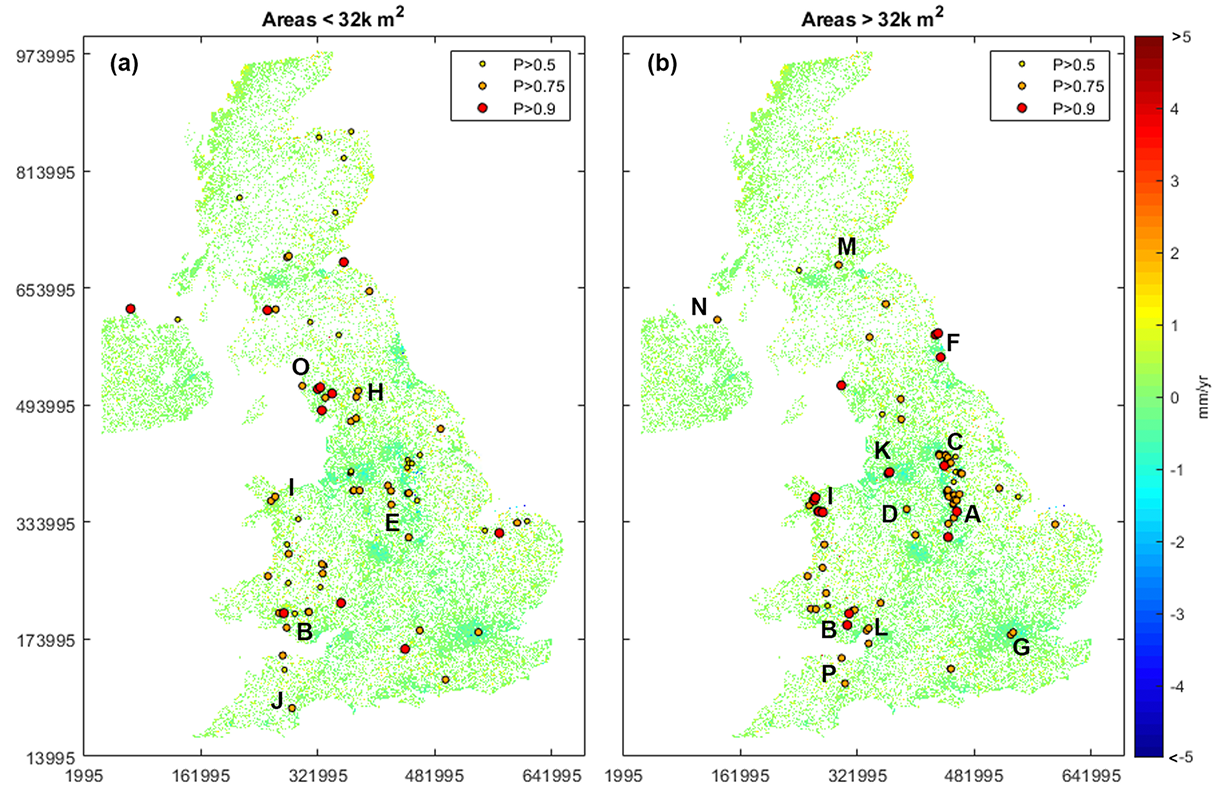}  
					\caption{Detection results in the 10-m interferograms showing the centre locations where the CNN identifies with high probability of deformation. For visualisation, the small and large areas are plotted separately (Left: the area size less than 32~km$^2$. Right: the area size larger than 32~km$^2$) }
    \label{fig:UK_map_resultsS}
\end{figure*} 

\begin{figure*}[h!]
	\centering
      	  \includegraphics[width=\textwidth]{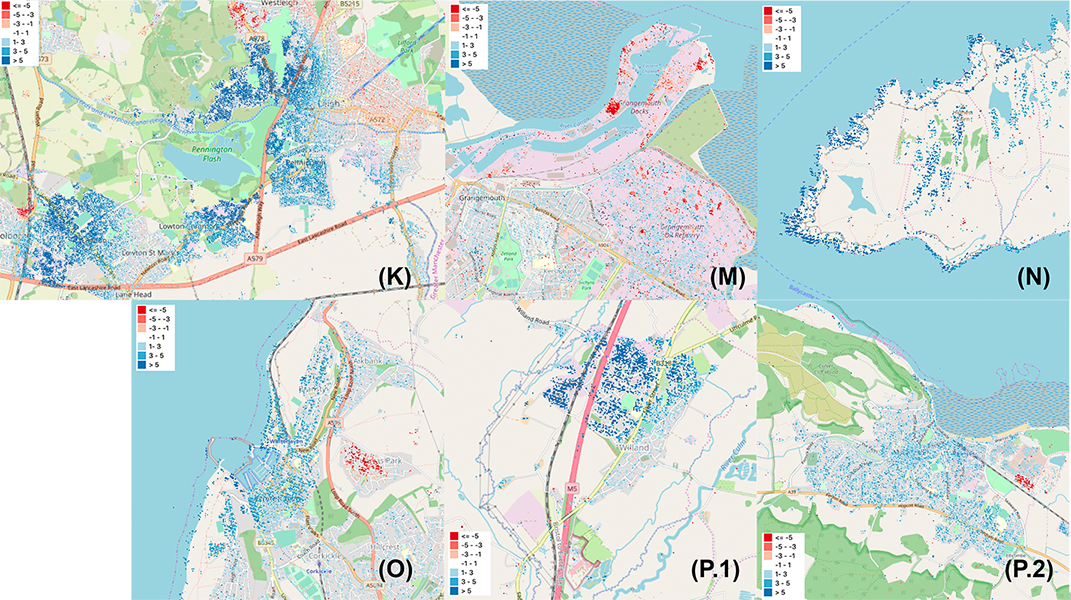}  
					\caption{Ground movement at area K (Golborne), area M (Grangemouth), area N (Rathlin Island), area O (Whitehaven), and area P (Willand - P1 and Minehead - P2). }
    \label{fig:AtoZ_results}
\end{figure*}

\begin{figure*}[h!]
	\centering
      	  \includegraphics[width=0.7\textwidth]{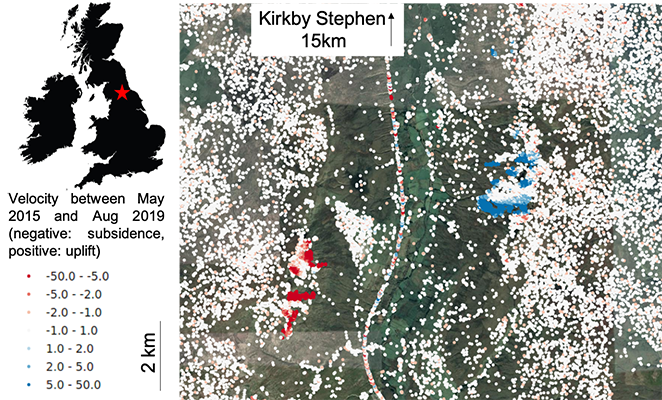}  
					\caption{Two landslides on two sides of a valley near Kirkby Stephen, one appears as subsidence and the other as uplift. This is the horizontal motion showing as  uplift and subsidence.}
    \label{fig:Kirkby_Stephen_132A}
\end{figure*}

\end{document}